\documentclass[runningheads]{llncs}
\usepackage{graphicx}
\usepackage{comment}
\usepackage{amsmath,amssymb} 
\usepackage{color}
\usepackage{cite}
\usepackage{booktabs}
\usepackage{caption}
\usepackage{hyperref}

\begin{document}
\pagestyle{headings}
\mainmatter
\def\ECCVSubNumber{277}  

\title{Atlas: End-to-End 3D Scene Reconstruction from Posed Images} 

\titlerunning{Atlas}
%

\author{Zak Murez\inst{1} \and
Tarrence van As\inst{2*} \and
James Bartolozzi \inst{*} \and 
Ayan Sinha\inst{1} \and
Vijay Badrinarayanan \inst{3*} \and
Andrew Rabinovich \inst{2*}}
\authorrunning{Z. Murez et al.}
%
\institute{Magic Leap Inc., CA, USA \email{zak@murez.com,asinha@magicleap.com}\\
*Work done at Magic Leap \email{bartolozzij@gmail.com} \and
InsideIQ Inc., CA, USA \email{\{tarrence,andrew\}@insideiq.team} \and
Wayve.ai, London, UK \email{vijay@wayve.ai} 
}

\maketitle

\begin{abstract}
We present an end-to-end 3D reconstruction method for a scene by directly regressing a truncated signed distance function (TSDF) from a set of posed RGB images. Traditional approaches to 3D reconstruction rely on an intermediate representation of depth maps prior to estimating a full 3D model of a scene. We hypothesize that a direct regression to 3D is more effective. A 2D CNN extracts features from each image independently which are then back-projected and accumulated into a voxel volume using the camera intrinsics and extrinsics. After accumulation, a 3D CNN refines the accumulated features and predicts the TSDF values. Additionally, semantic segmentation of the 3D model is obtained without significant computation. This approach is evaluated on the Scannet dataset where we significantly outperform state-of-the-art baselines (deep multiview stereo followed by traditional TSDF fusion) both quantitatively and qualitatively. We compare our 3D semantic segmentation to prior methods that use a depth sensor since no previous work attempts the problem with only RGB input.

\end{abstract}

\section{Introduction}
Reconstructing the world around us is a long standing goal of computer vision.
Recently many applications have emerged,
such as autonomous driving and augmented reality,
which rely heavily upon accurate 3D reconstructions of the surrounding environment.
These reconstructions are often estimated by fusing depth measurements from special sensors, such as structured light, time of flight, or LIDAR, into 3D models.
While these sensors can be extremely effective,
they require special hardware making them more cumbersome and expensive than systems that rely solely on RGB cameras.
Furthermore, they often suffer from noise and missing measurements due to low albedo and glossy surfaces as well as occlusion.

Another approach to 3D reconstruction is to use monocular \cite{lasinger2019towards, lee2019big, fu2018deep}, binocular \cite{chang2018pyramid, chabra2019stereodrnet} or multivew \cite{im2019dpsnet, huang2018deepmvs, wang2018mvdepthnet, hirschmuller2007stereo} stereo methods which take RGB images (one, two, or multiple respectively) and predict depth maps for the images.
Despite the plethora of recent research, these methods are still much less accurate than depth sensors, and
do not produce satisfactory results when fused into a 3D model.

\begin{figure}
\centering
\includegraphics[width=.99\linewidth]{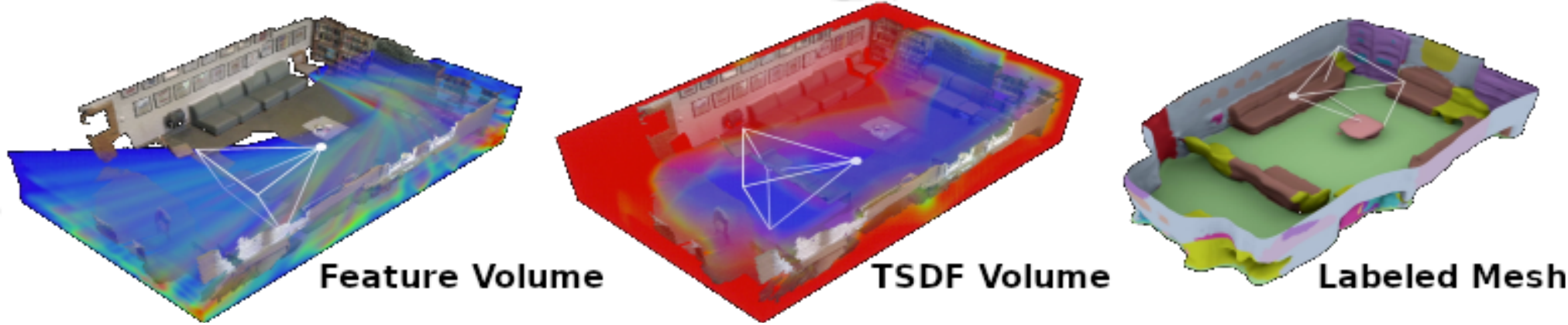}
\caption{
Overview of our method. Features from each image are backprojected along rays and accumulated into a feature volume.
Then a 3D CNN refines the features and regresses a TSDF volume. Finally a mesh is extracted from the TSDF.
Semantic Labels can also be output.
}

\label{fig:teaser}
\end{figure}

In this work, we observe that depth maps are often just intermediate representations that
are then fused with other depth maps into a full 3D model.
As such, we propose a method that takes a sequence of RGB images
and directly predicts a full 3D model in an end-to-end trainable manner.
This allows the network to fuse more information and learn better geometric priors about the world,
producing much better reconstructions.
Furthermore, it reduces the complexity of the system by eliminating steps like frame selection,
as well as reducing the required compute by amortizing the cost over the entire sequence.

Our method is inspired by two main lines of work: cost volume based multi view stereo \cite{im2019dpsnet, yao2018mvsnet} and Truncated Signed Distance Function (TSDF) refinement \cite{dai2019sg, dai2018scancomplete}.
Cost volume based multi view stereo methods construct a cost volume using a plane sweep.
Here, a reference image is warped onto the target image for each of a fixed set of depth planes and stacked into a 3D cost volume.
For the correct depth plane, the reference and target images will match while for other depth planes they will not.
As such, the depth is computed by taking the argmin over the planes.
This is made more robust by warping image features extracted by a CNN instead of the raw pixel measurements,
and by filtering the cost volume with another CNN prior to taking the argmin.

TSDF refinement starts by fusing depth maps from a depth sensor into an initial voxel volume using TSDF fusion \cite{curless1996volumetric},
in which each voxel stores the truncated signed distance to the nearest surface.
Note that a triangulated mesh can then be extracted from this implicit representation by finding the zero crossing surface using marching cubes \cite{lorensen1987marching}.
TSDF refinement methods \cite{dai2019sg, dai2018scancomplete} take this noisy, incomplete TSDF as input and refine it by passing it
through a 3D convolutional encoder-decoder network.

Similar to cost volume multi view stereo approaches, we start by using a 2D CNN to extract features from a sequence of RGB images.
These features are then back projected into a 3D volume using the known camera intrinsics and extrinsics.
However, unlike cost volume approaches which back project the features into a target view frustum using image warping,
we back project into a canonical voxel volume, where each pixel gets mapped to a ray in the volume (similar to \cite{sitzmann2019deepvoxels}).
This avoids the need to choose a target image and allows us to fuse an entire sequence of frames into a single volume.
We fuse all the frames into the volume using a simple running average.
Next, as in both cost volume and TSDF refinement, we pass our voxel volume through a 3D convolutional encoder-decoder
to refine the features.
Finally, as in TSDF refinement, our feature volume is used to regress the TSDF values at each voxel (see Figure \ref{fig:teaser}).

We train and evaluate our network on real scans of indoor rooms from the Scannet\cite{dai2017scannet} dataset.
Our method significantly outperforms state-of-the-art multi view stereo baselines \cite{im2019dpsnet, wang2018mvdepthnet} producing
accurate and complete meshes.

As an additional bonus, for minimal extra compute, we can add an additional head to our 3D CNN and perform 3D semantic segmentation.
While the problems of 3D semantic and instance segmentation have received a lot of attention recently \cite{hou20183dsis, graham20183d},
all previous methods assume the depth was acquired using a depth sensor.
Although our 3D segmentations are not competitive with the top performers on the Scannet benchmark leader board,
we establish a strong baseline for the new task of 3D semantic segmentation from multi view RGB.


\section{Related Work}
\subsection{3D reconstruction}
Reconstructing a 3D model of a scene usually involves acquiring depth for a sequence of images and fusing the depth maps using a 3D data structure.
The most common 3D structure for depth accumulation is the voxel volume used by TSDF fusion\cite{curless1996volumetric}.
However, surfels (oriented point clouds) are starting to gain popularity \cite{whelan2015elasticfusion, schops2019surfelmeshing}.
These methods are usually used with a depth sensor, but can also be applied to depth maps predicted from monocular or stereo images.

With the rise of deep learning, monocular depth estimation has seen huge improvements \cite{lasinger2019towards, lee2019big, fu2018deep},
however their accuracy is still far below state-of-the-art stereo methods.
A popular classical approach to stereo \cite{hirschmuller2007stereo} uses mutual information and semi global matching to compute the disparity between two images. Similar approaches have been incorporated into SLAM systems such as COLMAP \cite{schoenberger2016mvs, schoenberger2016sfm} and CNN-SLAM\cite {tateno2017cnn}.
More recently, several end-to-end plane sweep algorithms have been proposed.
DeepMVS\cite{huang2018deepmvs} uses a patch matching network.
MVDepthNet\cite{wang2018mvdepthnet} constructs the cost volume from raw pixel measurements and performs 2D convolutions,
treating the planes as feature channels.
GPMVS\cite{hou2019multi} builds upon this and aggregates information into the cost volume over long sequences using a Gaussian process.
MVSNet\cite{yao2018mvsnet} and DPSNet\cite{im2019dpsnet} construct the cost volume from features extracted from the images using a 2D CNN. They then filter the cost volume using 3D convolutions on the 4D tensor.
R-MVSNet\cite{yao2019recurrent} reduces the memory requirements of MVSNet by replacing the 3D CNN with a recurrent CNN, while P-MVSNet\cite{chen2019point} starts with a low resolution MVSNet and then iteratively refines the estimate using their point flow module.
All of these methods require choosing a target image to predict depth for and then finding suitable neighboring reference images.
Recent binocular stereo methods \cite{chang2018pyramid, chabra2019stereodrnet} use a similar cost volume approach,
but avoid frame selection by using a fixed baseline stereo pair.
Depth maps over a sequence are computed independently (or weakly coupled in the case of \cite{hou2019multi}).
In contrast to these approaches, our method constructs a single coherent 3D model from a sequence of input images directly.

While TSDF fusion is simple and effective, it cannot reconstruct partially occluded geometry and requires averaging many measurements to reduce noise.
As such, learned methods have been proposed to improve the fusion.
OctNet-Fusion\cite{riegler2017octnet} uses a 3D encoder-decoder to aggregate multiple depth maps into a TSDF and shows results on single objects and portions of scans.
ScanComplete\cite{dai2018scancomplete} builds upon this and shows results for entire rooms.
SG-NN\cite{dai2019sg} improves upon ScanComplete by increasing the resolution using sparse convolutions\cite{graham20183d} and training using a novel self-supervised training scheme.
3D-SIC\cite{hou20193d} focuses on 3D instance segmentation using region proposals and adds a per instance completion head.
Routed fusion\cite{weder2020routedfusion} uses 2D filtering and 3D convolutions in view frustums to improve aggregation of depth maps.

More similar in spirit to ours are networks that take one or more images and directly predict a 3D representation.
3D-R2N2 \cite{choy20163d} encodes images to a latent space and then decodes a voxel occupancy volume.
Octtree-Gen\cite{tatarchenko2017octree} increases the resolution by using an octtree data structure to improve the efficiency of 3D voxel volumes.
Deep SDF\cite{park2019deepsdf} chooses to learn a generative model that can output an SDF value for any input position instead of discretizing the volume.
These methods encode the input to a small latent code and report results on single objects, mostly from shapenet\cite{chang2015shapenet}.
This small latent code is unlikely to contain enough information to be able to reconstruct an entire scene (follow up work \cite{chabra2020deep}, concurrent with ours, addresses this problem, but they do not apply it to RGB only reconstruction).
Pix2Vox \cite{xie2019pix2vox} encodes each image to a latent code and then decodes a voxel representation for each and then fuses them. This is similar to ours, but we explicitly model the 3D geometry of camera rays allowing us to learn better representations and scale to full scenes.
SurfNet \cite{sinha2017surfnet} learns a 3D offset from a template UV map of a surface.
Point set generating networks\cite{fan2017point} learns to generate point clouds with a fixed number of points.
Pixel2Mesh++\cite{wang2018pixel2mesh} uses a graph convolutional network to directly predict a triangulated mesh.
Mesh-RCNN \cite{gkioxari2019mesh} builds upon 2D object detection\cite{he2017mask} and adds an additional head to predict a voxel occupancy grid for each instance and then refines them using a graph convolutional network on a mesh.

Back projecting image features into a voxel volume and then refining them using a 3D CNN has also been used for human pose estimation \cite{zimmermann2019freihand, iskakov2019learnable}.
These works regress 3D heat maps that are used to localize joint locations.

Deep Voxels \cite{sitzmann2019deepvoxels} and the follow up work of scene representation networks\cite{sitzmann2019srns} accumulate features into a 3D volume forming an unsupervised representation of the world which can then be used to render novel views without the need to form explicit geometric intermediate representations.

\subsection{3D Semantic Segmentation}
In addition to reconstructing geometry, many applications require semantic labeling of the reconstruction to provide a richer representation. Broadly speaking, there are two approaches to solving this problem: 1) Predict semantics on 2D input images using a 2D segmentation network\cite{ badrinarayanan2015segnet,he2017mask,cheng2019panoptic} and back project the labels to 3D \cite{mccormac2017semanticfusion, narita2019panopticfusion, mccormac2018fusion++} 2) Directly predict the semantic labels in the 3D space.
All of these methods assume depth is provided by a depth sensor.
A notable exception is Kimera \cite{Rosinol20Kimera}, which uses multiview stereo \cite{hirschmuller2007stereo} to predict depth,
however, they only show results on synthetic data and ground truth 2D segmentations.

SGPN\cite{wang2017sgpn} formulates instance segmentation as a 3D point cloud clustering problem. Predicting a similarity matrix and clustering the 3D point cloud to derive semantic and instance labels. 3D-SIS\cite{hou20183dsis} improves upon these approaches by fusing 2D features in a 3D representation. RGB images are encoded using a 2D CNN and back projected onto the 3D geometry reconstructed from depth maps. A 3D CNN is then used to predict 3D object bounding boxes and semantic labels.
SSCN \cite{graham20183d} predicts semantics on a high resolution voxel volume enabled by sparse convolutions.

In contrast to these approaches, we propose a strong baseline to the relatively untouched problem of 3D semantic segmentation without a depth sensor.



\section{Method}
\begin{figure}
\centering
\includegraphics[width=.99\linewidth]{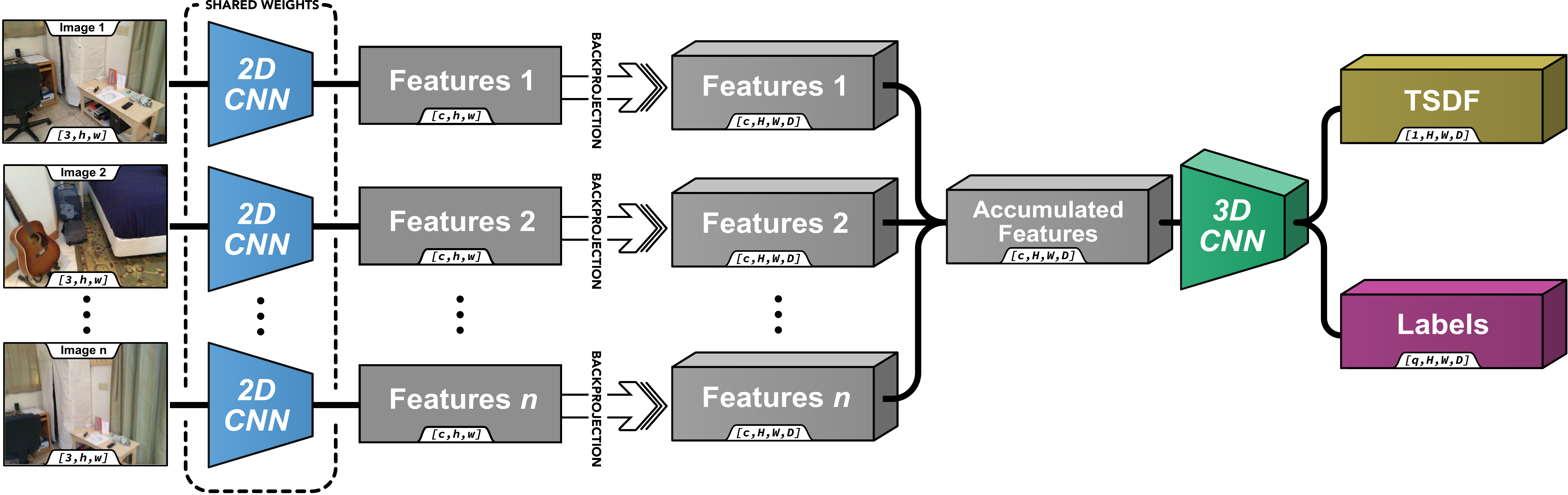}
\caption{Schematic of our method. Features are extracted from a sequence of images using a 2D CNN and then back projected into a 3D volume.
These volumes are accumulated and then passed through a 3D CNN to directly regress a TSDF reconstruction of the scene.
We can also jointly predict the 3D semantic segmentation of the scene.
}
\label{fig:overview}
\end{figure}

Our method takes as input an arbitrary length sequence of RGB images, each with known intrinsics and pose.
These images are passed through a 2D CNN backbone to extract features.
The features are then back projected into a 3D voxel volume and accumulated using a running average.
Once the image features have been fused into 3D, we regress a TSDF directly using a 3D CNN (See Fig.~\ref{fig:overview}).
We also experiment with adding an additional head to predict semantic segmentation.



\subsection{Feature Volume Construction}
Let $I_t \in \mathbb{R}^{3 \times h \times w}$ be an image in a sequence of $T$ RGB images.
We extract features $F_t = F(I_t) \in \mathbb{R}^{c \times h \times w}$ using a standard 2D CNN where $c$ is the feature dimension.
These 2D features are then back projected into a 3D voxel volume using the known camera intrinsics and extrinsics,
assuming a pinhole camera model. Consider a voxel volume $V \in \mathbb{R}^{c \times H \times W \times D}$
\begin{equation}
    V_t(:,i,j,k) = F_t(:,\hat{i},\hat{j}), \hspace{0.2cm}\mathrm{with} 
\end{equation}
\begin{equation}
    \begin{bmatrix}
        \hat{i} \\
        \hat{j}
    \end{bmatrix}
    =
    \Pi K_t P_t
    \begin{bmatrix}
        i \\
        j \\
        k \\
        1
    \end{bmatrix},
\end{equation}
where $P_t$ and $K_t$ are the extrinsics and intrinsics matrices for image $t$ respectively, $\Pi$ is the perspective mapping and $:$ is the slice operator.
Here $(i,j,k)$ are the voxel coordinates in world space and $(\hat{i},\hat{j})$ are the pixel coordinates in image space.
Note that this means that all voxels along a camera ray are filled with the same features corresponding to that pixel.

These feature volumes are accumulated over the entire sequence using a weighted running average similar to TSDF fusion as follows:
\begin{equation}
    \bar{V}_t = \frac{ \bar{V}_{t-1} \bar{W}_{t-1} + V_t }
                     { \bar{W}_{t-1} + W_t},
\end{equation}
\begin{equation}    
    \bar{W}_t = \bar{W}_{t-1} + W_t.
\end{equation}
For the weights we use a binary mask $W_t(i,j,k) \in \{0, 1\}$ which stores if voxel $(i,j,k)$ is inside or outside the view frustum of the camera.






\subsection{3D Encoder-Decoder}
\begin{figure}
\centering
\includegraphics[width=.6\linewidth]{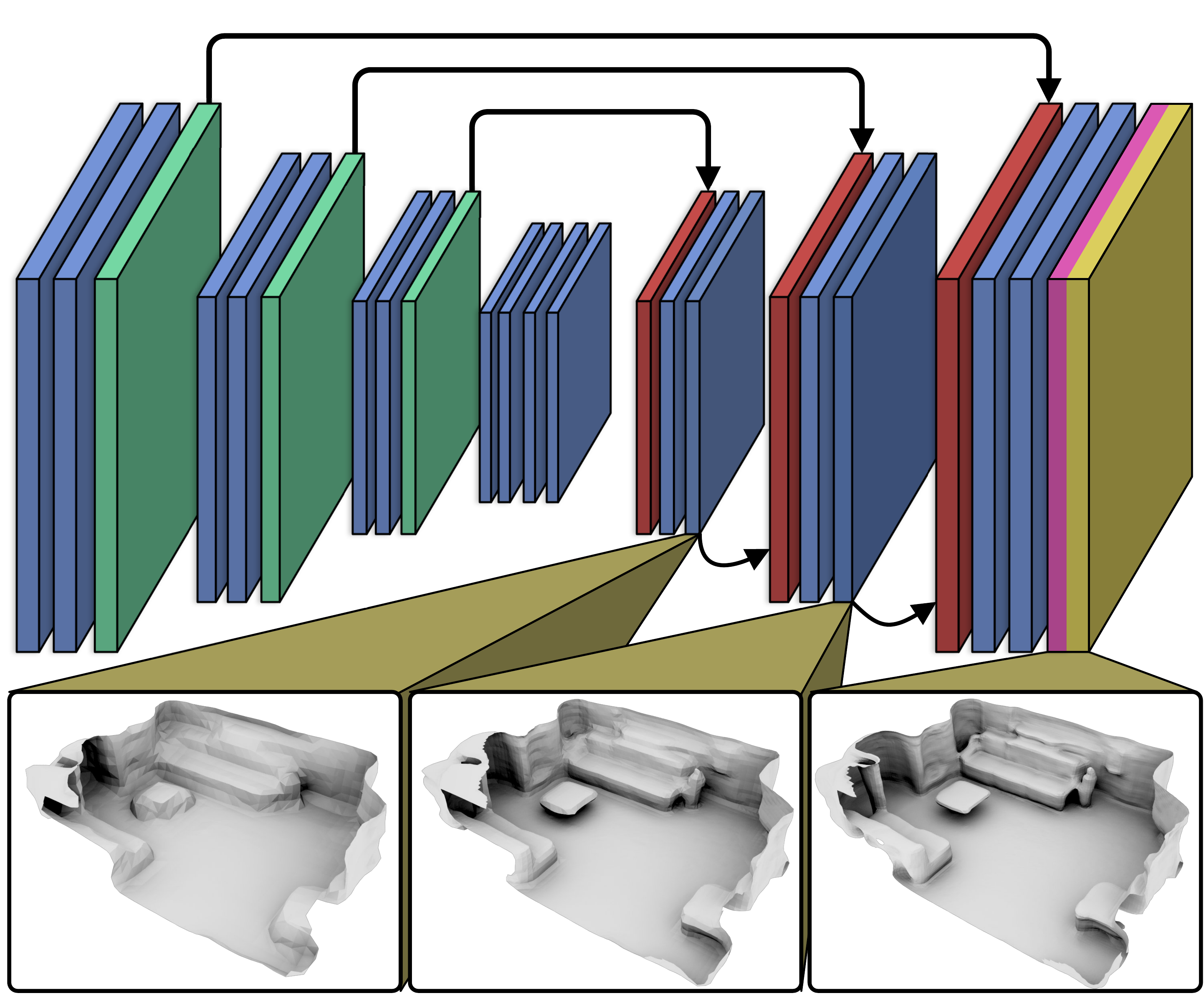}
\caption{Our 3D encoder-decoder architecture. 
Blue boxes denote residual blocks,
green boxes are stride 2 convolutions and red boxes are trilinear upsampling.
The arrows from the encoder to the decoder indicate skip connections.
Our network predicts TSDFs in a coarse to fine manner with the previous resolution being used to sparsify the next resolution (shown as small arrows in the decoder).
}
\label{fig:3dcnn}
\end{figure}

Once the features are accumulated into the voxel volume, we use a 3D convolutional encoder-decoder network
to refine the features and regress the output TSDF (Fig.~\ref{fig:3dcnn}).
Each layer of the encoder and decoder uses a set of 3x3x3 residual blocks.
Downsampling is implemented with 3x3x3 stride 2 convolution,
while upsampling uses trilinear interpolation followed by a 1x1x1 convolution to change the feature dimension.
The feature dimension is doubled with each downsampling and halved with each upsampling.
All convolution layers are followed by batchnorm and relu.
We also include additive skip connections from the encoder to the decoder.

At the topmost layer of the encoder-decoder, we use a 1x1x1 convolution followed by a tanh activation to regress the final TSDF values.
For our semantic segmentation models we also include an additional 1x1x1 convolution to predict the segmentation logits.

We also include intermediate output heads at each decoded resolution prior to upsampling.
These additional predictions are used both for intermediate supervision to help the network train faster,
as well as to guide the later resolutions to focus on refining predictions near surfaces.
At each resolution, any voxel that is predicted beyond a fraction (.99) of the truncation distance
is clamped to one at the following resolutions.
Furthermore, loss is only backpropageted for non-clamped voxels.
Without this, the loss at the higher resolutions is dominated by the large number of empty space voxels
and the network has a harder time learning fine details.

Note that since our features are back projected along entire rays, the voxel volume is filled densely and thus we cannot take advantage of sparse convolutions\cite{graham20183d} in the encoder.
However, the multiscale outputs can be used to sparsify the feature volumes in the decoder allowing for the use of sparse convolutions similar to \cite{dai2019sg}.
In practice, we found that we were able to train our models at $4cm^3$ voxel resolution without the need for sparse convolutions.

\section{Implementation Details}


We use a Resnet50-FPN\cite{lin2017feature} followed by the merging method of \cite{kirillov2019panoptic} with 32 output feature channels as our 2D backbone.
Our 3D CNN consists of a four scale resolution pyramid where we double the number of channels each time we half the resolution.
The encoder consists of (1,2,3,4) residual blocks at each scale respectively, and the decoder consists of (3,2,1) residual blocks.

We supervise the multiscale TSDF reconstructions using $\ell_1$ loss to the ground truth TSDF values.
Following \cite{dai2016shape}, we log-transform the predicted and target values before applying the $\ell_1$ loss, 
and only backpropagate loss for voxels that were observed in the ground truth (i.e. have TSDF values strictly less than $1$.)
However, to prevent the network from hallucinating artifacts behind walls, outside the room, we also mark all the voxels where their entire vertical column is equal to $1$ and penalize in these areas too.
The intuition for this is that if the entire vertical column was not observed it was probably not within the room.
To construct the ground truth TSDFs we run TSDF fusion at each resolution on the full sequences, prior to training.

We train the network end-to-end using 50 images selected randomly throughout the full sequence.
We use a voxel size of $4cm^3$ with a grid of $(160\times160\times64)$ voxels, corresponding to a volume of $(6.4\times 6.4 \times 2.56)$ meters.
At test time, we accumulate the feature volumes in place (since we do not need to store the intermediate activations for backpropagation), allowing us to operate on arbitrary length sequences (often thousands of frames for ScanNet) and we use a 400x400x104 sized voxel grid corresponding to a volume of $(16\times16\times4.16)$ meters. 
We use the ADAM optimizer with a learning rate of $5\mathrm{e}{-4}$ and 16bit mixed precision operations.
Training the network takes around 24 hours on 8 Titan RTX GPUs with a batch size of 8 (1 sequence per GPU) and synchronized batchnorm. Our model is implemented with PyTorch and PyTorch Lightning\cite{falcon2019pytorch}.



 
\section{Results}

\begin{figure}
\centering
\includegraphics[width=.75\linewidth]{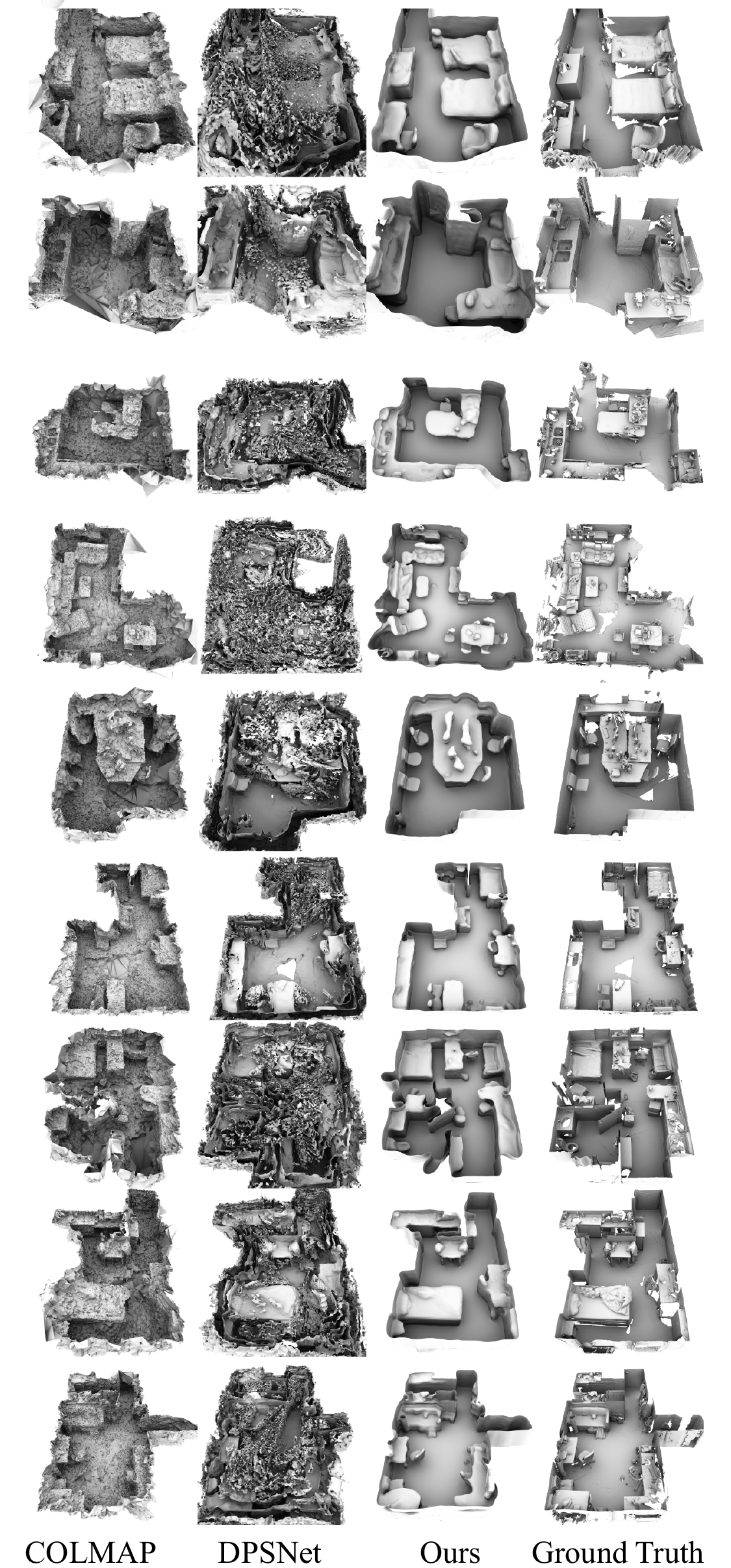}
\caption{Qualitative 3D reconstruction results.
}
\label{fig:geo1}
\end{figure}

We evaluate our method on ScanNet\cite{dai2017scannet}, which consists of 2.5M images across 707 distinct spaces.
Standard train/validation/test splits are adopted. The 3D reconstructions are benchmarked using standard 2D depth metrics (Table \ref{tab:results_depth}) and 3D metrics (Table \ref{tab:results_geo}), which are defined in Table \ref{tab:metric_defs}. We also show qualitative comparisons in Figure \ref{fig:geo1} where our method really stands out.




We compare our method to 4 state-of-the-art baselines: COLMAP \cite{schoenberger2016mvs, schoenberger2016sfm}, MVDepthNet\cite{wang2018mvdepthnet}, GPMVS\cite{hou2019multi}, and DPSNet\cite{im2019dpsnet}.
For COLMAP we use the default dense reconstruction parameters but use the ground truth poses provided by Scannet.
For each of the learned methods we fine tuned the models provided by the authors on Scannet. At inference time, 6 reference frames were selected temporally with stride 10 centered around the target view. We also mask the boundary pixels since the networks have visible edge effects that cause poor depth predictions here (leading to 92.8\% completeness).

To evaluate these in 3D we fuse the predicted depth maps using two techniques: TSDF Fusion \cite{curless1996volumetric} and point cloud fusion.
For COLMAP we use their default point cloud fusion, while for the other methods we use the implementation of \cite{galliani2015massively}. We found point cloud fusion was more robust to the outliers present in the depth predictions than our implementation of TSDF Fusion. As such, we only report the point cloud fusion results in Table \ref{tab:results_geo} which are strictly better than the TSDF Fusion results (Note that the $L_1$ metric is computed using the TSDF Fusion approach as it is not computed in the point cloud fusion approach).

\begin{figure}
\centering
\includegraphics[width=.85\linewidth]{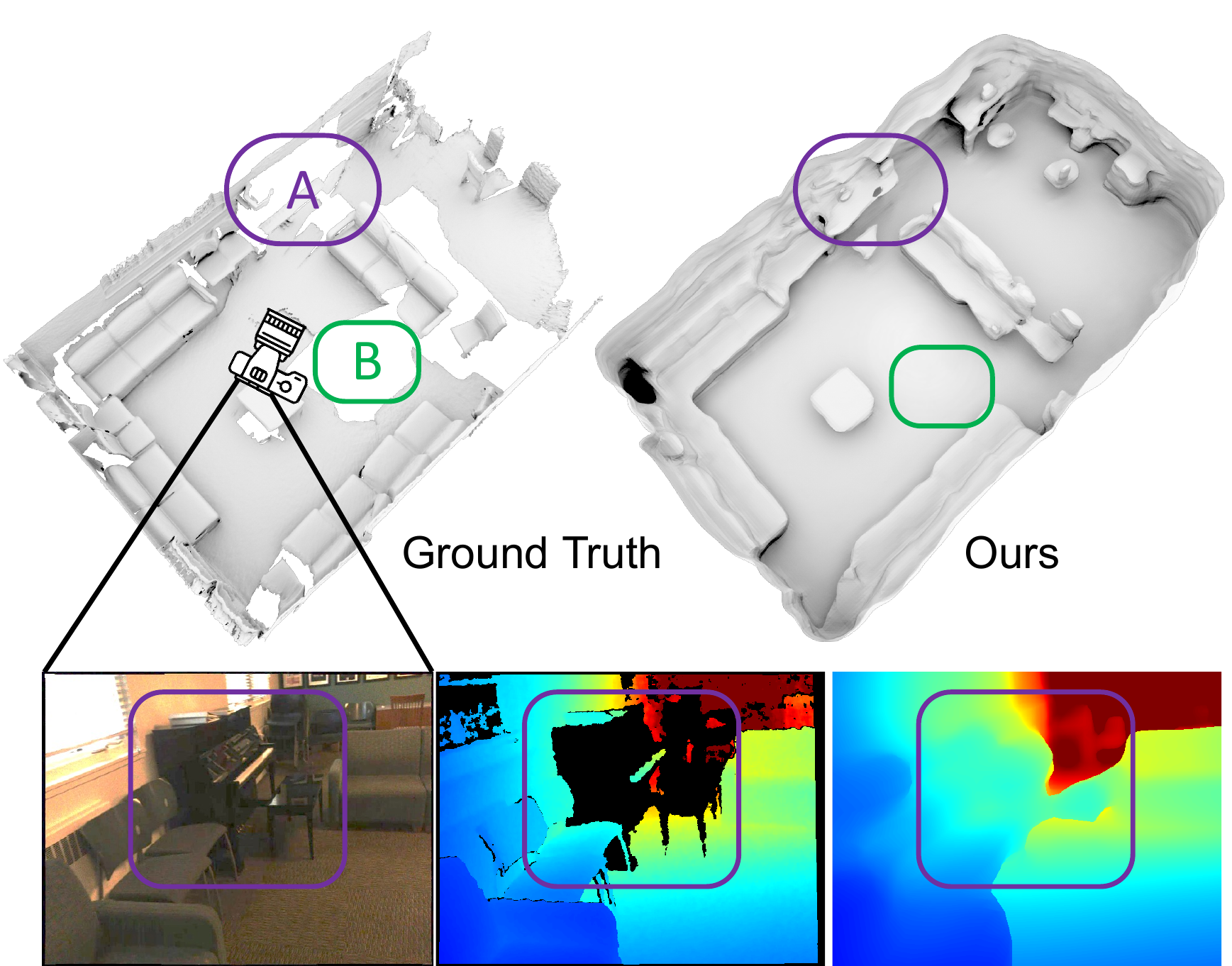}
\caption{Our method learns to fill holes that are missing from the ground truth.
These holes arise from two causes: A) limitations of depth sensors on low albedo and specular surfaces, and B) unobserved regions caused by occlusion and incomplete scans. While other multiview stereo method often learn to predict depth for these troublesome surfaces, they are not able to complete unobserved geometry.
}
\label{fig:fill}
\end{figure}

As seen in Figure~\ref{fig:fill} our method is able to fill holes that are missing from the ground truth.
These holes arise from two causes: A) limitations of depth sensors on low albedo and specular surfaces, and B) unobserved regions caused by occlusion and incomplete scans. While other multiview stereo method often learn to predict depth for these troublesome surfaces, they are not able to complete unobserved geometry. On the other hand, since our method directly regresses the full TSDF for a scene, it is able to reason about and complete unobserved regions. However, this means that we must take extra care when evaluating the point cloud metrics, otherwise we will be falsely penalized in these regions. We remove geometry that was not observed in the ground truth by taking the rendered depth maps from our predicted mesh and re-fuse them using TSDF Fusion into a trimmed mesh. This guarantees that there is no mesh in areas that were not observed in the ground truth.

Our method achieves state-of-the-art on about half of the metrics and is competitive on all metrics. However, as seen in Figure~\ref{fig:geo1}, qualitatively our results our significantly better than previous methods. While the $L_1$ metric on the TSDF seems to reflect this performance gap better, the inability of the other metrics to capture this indicates a need for additional more perceptual metrics.

\begin{figure}
\centering
\includegraphics[width=.9\linewidth]{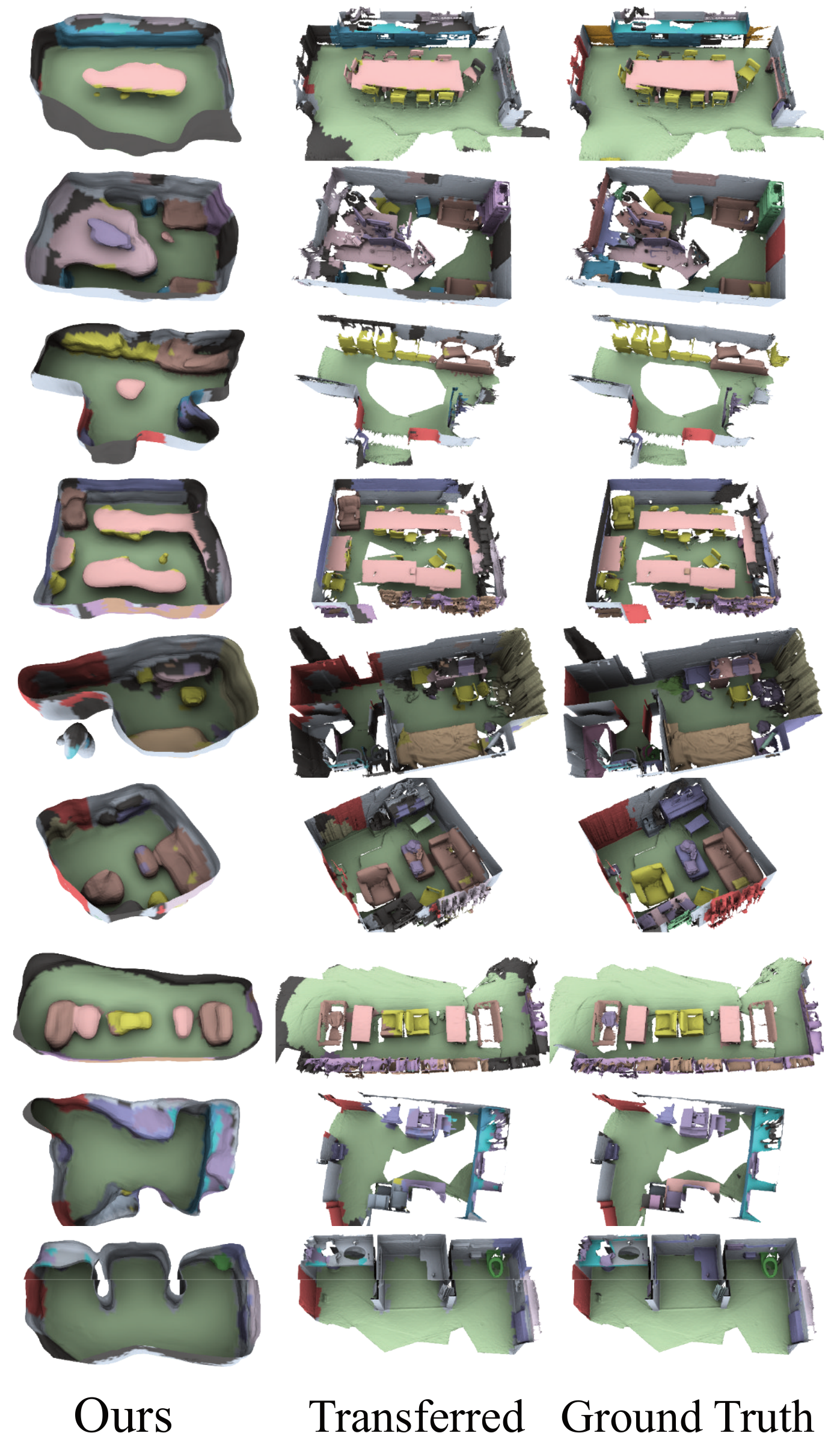}
\caption{Qualitative 3D semantic segmentations. Left to right: Ours, our labels transferred to the ground truth mesh, ground truth labels.
We are able to accurately segment the 3D scene despite not using a depth sensor.
}
\label{fig:semseg}
\end{figure}

As mentioned previously, we augment the existing 3D-CNN with a semantic segmentation head, requiring only a single $1\times1\times1$ convolution, to be able to not only reconstruct the 3D structure of the scene but also provide semantic labels to the surfaces. Since no prior work attempts to do 3D semantic segmentation from only RGB images, and there are no established benchmarks, we propose a new evaluation procedure. The semantic labels from the predicted mesh are transferred onto the ground truth mesh using nearest neighbor lookup on the vertices, and then the standard IOU metric can be used. The results are reported in Table \ref{tab:scannet} and Fig. \ref{fig:semseg} (note that this is an unfair comparison since all prior methods include depth as input).

\begin{table}[]
\centering
\caption{Definitions of metrics: $n$ is the number of pixels with both valid ground truth and predictions, $d$ and $d^*$ are the predicted and ground truth depths (the predicted depth from our method is computed by rendering the predicted mesh). $t$ and $t^*$ are the predicted and ground truth TSDFs while $p$ and $p^*$ are the predicted and ground truth point clouds.
}
\begin{tabular}{llll}
\hline
\multicolumn{2}{c}{2D} & \multicolumn{2}{c}{3D} \\
\hline
Abs Rel  & $\frac{1}{n}\sum{|d-d^*|/d^*}$ & L1 & $\mbox{mean}_{t^*<1}{|t-t^*|}$\\
Abs Diff & $\frac{1}{n}\sum{|d-d^*|}$ & Acc & $\mbox{mean}_{p \in P}(\min_{p^*\in P^*}||p-p^*||)$ \\
Sq Rel   & $\frac{1}{n}\sum{|d-d^*|^2/d^*}$ & Comp & $\mbox{mean}_{p^* \in P^*}(\min_{p\in P}||p-p^*||)$ \\
RMSE     & $\sqrt{\frac{1}{n}\sum{|d-d^*|^2}}$ & Prec & $\mbox{mean}_{p \in P}(\min_{p^*\in P^*}||p-p^*||<.05)$ \\
$\delta < 1.25^i$ & $\frac{1}{n}\sum{(\max{(\frac{d}{d^*},\frac{d^*}{d})} < 1.25^i)}$ & Recal & $\mbox{mean}_{p^* \in P^*}(\min_{p\in P}||p-p^*||<.05)$ \\
Comp & $\%$ valid predictions & F-score & $\frac{ 2 \times \text{Perc} \times \text{Recal} }{\text{Perc} + \text{Recal}}$ \\
\hline
\end{tabular}
\label{tab:metric_defs}
\end{table}

\begin{table}[]
\centering
\caption{2D Depth Metrics
}
\begin{tabular}{lcccccccc}
\hline
Method & AbsRel & AbsDiff & SqRel & RMSE & $\delta<1.25$ & $\delta<1.25^2$ & $\delta<1.25^3$ & Comp\\
\hline
COLMAP \cite{schoenberger2016mvs}    & .137 & .264 & .138 & .502 & .834 & .908 & .938 & .871 \\
MVDepthNet \cite{wang2018mvdepthnet} & .098 & .191 & .061 & .293 & .896 & .977 & .994 & .928 \\
GPMVS \cite{hou2019multi}            & .130 & .239 & .339 & .472 & .906 & .967 & .980 & .928 \\
DPSNet \cite{im2019dpsnet}           & .087 & .158 & \textbf{.035} & \textbf{.232} & .925 & \textbf{.984} & \textbf{.995} & .928 \\
\midrule
Ours (plain)                         & \textbf{.061} & \textbf{.120} & .042 & .248 & \textbf{.940} & .972 & .985 &  \textbf{.999} \\
Ours (semseg)                        & .065 & .124 & .043 & .251 & .936 & .971 & .986 & \textbf{.999} \\
\hline

\end{tabular}

\label{tab:results_depth}
\end{table}

\begin{table}[]
\centering
\caption{3D Geometry Metrics
}
\begin{tabular}{lcccccc}
\hline
Method & $L_1$ & Acc & Comp & Prec & Recal & F-score \\
\hline
COLMAP \cite{schoenberger2016mvs}    & .599 & .135 & .069 & \textbf{.505} & .634 & \textbf{.558} \\
MVDepthNet \cite{wang2018mvdepthnet} & .518 & .240 & .040 & .208 & .831 & .329 \\
GPMVS \cite{hou2019multi}            & .475 & .879 & \textbf{.031} & .188 & \textbf{.871} & .304 \\
DPSNet \cite{im2019dpsnet}           & .421 & .284 & .045 & .223 & .793 & .344 \\
\midrule
Ours (plain)                         & \textbf{.162} & .130 & .065 & .383 & .725 & .499 \\
Ours (semseg)                        & .172 & \textbf{.124} & .074 & .413 & .711 & .520 \\
\hline
\end{tabular}
\label{tab:results_geo}
\end{table}

From the results in Table \ref{tab:scannet} we see that our approach is surprisingly competitive with (and even beats some) prior methods that include depth as input. Having depth as an input makes the problem significantly easier because the only source of error is from the semantic predictions. In our case, in order to correctly label a vertex we must both predict the geometry correct as well as the semantic label. From Fig. \ref{fig:semseg} we can see that mistakes in geometry compounds with mistakes in semantics which leads to lower IOUs.



 \tabcolsep= 1mm
 \begin{table}[t]
     \centering
     \caption{3D Semantic Label Benchmark}
     \begin{tabular}{rc}
     \hline
             \small{Method}  & \small{mIOU} \\
    \hline
        ScanNet~\cite{dai2017scannet}  & 30.6\\
        PointNet++~\cite{qi2017pointnet} & 33.9\\
        SPLATNet~\cite{su2018splatnet} & 39.3 \\
        3DMV~\cite{dai20183dmv} & 48.4 \\
        3DMV-FTSDF & 50.1 \\
        PointNet++SW & 52.3 \\
        SparseConvNet~\cite{graham20183d} & 72.5 \\
        MinkowskiNet~\cite{choy20194d} & \textbf{73.4} \\
    \hline
        Ours & 34.0 \\
    \hline
    \end{tabular}
    \label{tab:scannet}
    \caption*{\small{ScanNet 3D Semantic Segmentation metrics. We transfer our labels from the predicted mesh to the ground truth mesh using nearest neighbors.}}
\end{table}

\begin{figure}
\centering
\includegraphics[width=.55\linewidth]{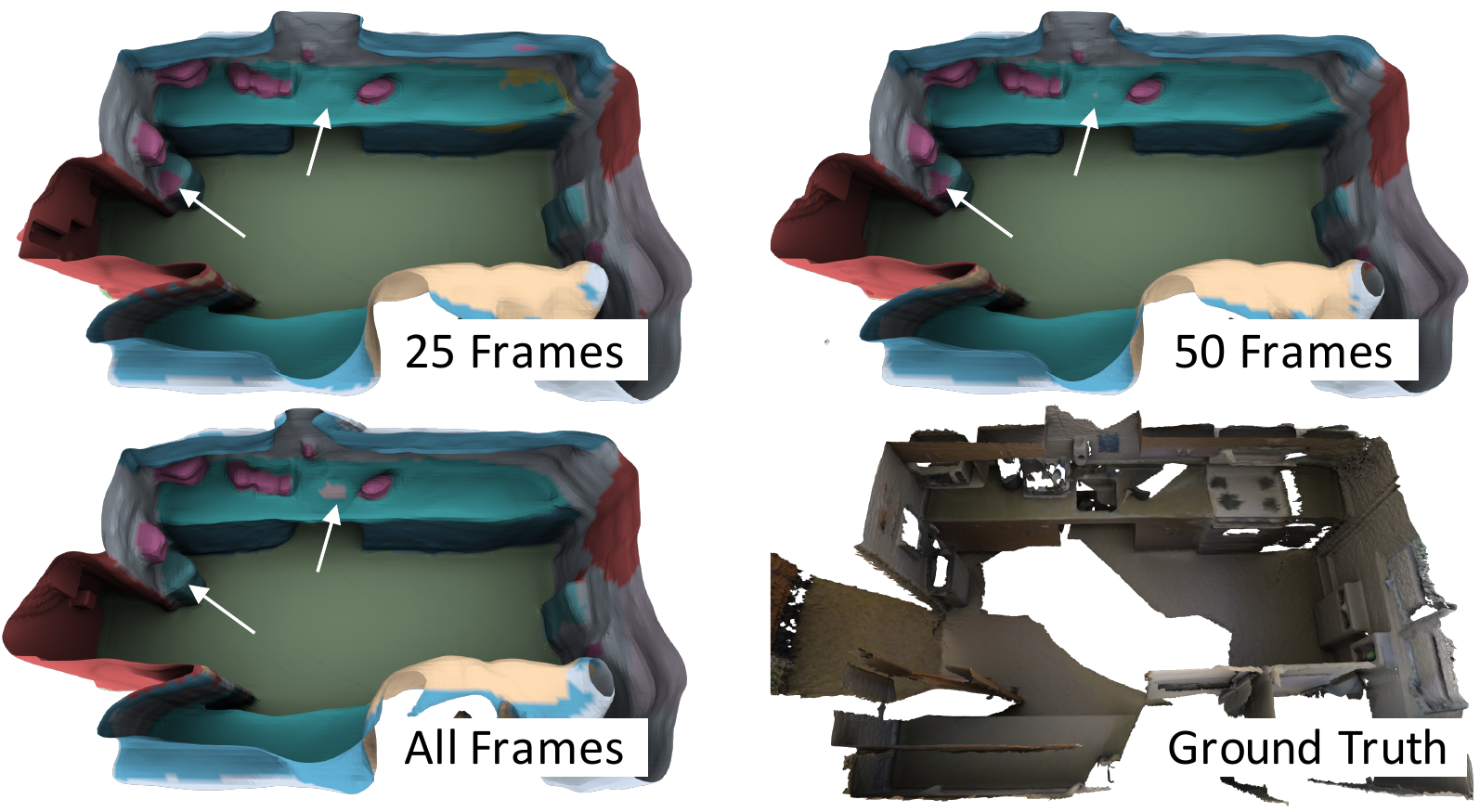}
\caption{Quality as a function of number of input frames at inference time.
There is almost no degradation with as few as 25 frames (out of 784 total).
}
\label{fig:ablation}
\end{figure}

In Figure \ref{fig:ablation} we show an example of how our method degrades as the number of frames is reduced at inference time. We see that there is almost no degradation with as few as 25 frames. See accompanying video for more examples.

\subsection{Inference Time}
Since our method only requires running a small 2D CNN on each frame, the cost of running the large 3D CNN is amortized over a sequence of images.
On the other hand, MVS methods must run all their compute on every frame.
Note that they must also run depth map fusion to accumulate the depth maps into a mesh, but we do not include this additional time here.
We report inference times using 2 neighbors.
All models are run on a single NVidia TiTan RTX GPU.
From Table \ref{tab:time} we can see that after approximately 4 frames, ours becomes faster than DPSNet (note that most Scannet scenes are a few thousands of frames).

\begin{table}[]
\centering
\caption{Inference Time}
\begin{tabular}{lcc}
\hline
Method & Per Frame Time (sec) & Per Sequence Time (sec)\\
\hline
COLMAP \cite{schoenberger2016mvs} & 2.076 & 0 \\
MVDepthNet \cite{wang2018mvdepthnet}& 0.048 & 0 \\
GPMVS \cite{hou2019multi} & 0.051 & 0 \\
DPSNet \cite{im2019dpsnet} & 0.322  & 0 \\
\midrule
Ours & .071 & .840 \\
\hline

\end{tabular}
\label{tab:time}
\end{table}

\section{Conclusions}
In this work, we present a novel approach to 3D scene reconstruction. Notably, our approach does not require depth inputs; is unbounded temporally, allowing the integration of long frame sequences; completes unobserved geometry; and supports the efficient prediction of other quantities such as semantics. We have experimentally verified that the classical approach to 3D reconstruction via per view depth estimation is inferior to direct regression to a 3D model from an input RGB sequence. We have also demonstrated that without significant additional compute, a semantic segmentation objective can be added to the model to accurately label the resultant surfaces. 
In our future work, we aim to improve the back projection and accumulation process. One approach is to allow the network to learn where along a ray to place the features (instead of uniformly). This will improve the models ability to handle occlusions and large multi room scenes.
We also plan to add additional tasks such as instance segmentation and intrinsic image decomposition.
Our method is particularly well suited for intrinsic image decomposition because the network has the ability to reason with information from multiple views in 3D.

%
%

\bibliographystyle{splncs04}
\bibliography{egbib}
\end{document}